\pdfoutput=1
\documentclass[11pt,a4paper]{article}
\usepackage[hyperref]{eacl2021}
\usepackage{times}

\usepackage{latexsym}

\usepackage{array}
\usepackage{tablefootnote}
\usepackage{graphicx}
\usepackage{multirow}
\usepackage{microtype}

\aclfinalcopy % Uncomment this line for the final submission
\usepackage{booktabs}

\title{ParaSCI: A Large Scientific Paraphrase Dataset for Longer Paraphrase Generation}
\author{Qingxiu Dong\textsuperscript{\rm 1,\rm 3,\rm 4}, Xiaojun Wan\textsuperscript{\rm 1,\rm 2,\rm 3} \and Yue Cao\textsuperscript{\rm 1,\rm 2,\rm 3} \\ 
  \textsuperscript{\rm 1}Wangxuan Institute of Computer Technology, Peking University\\ \textsuperscript{\rm 2}Center for Data Science, Peking University\\
  \textsuperscript{\rm 3}The MOE Key Laboratory of Computational Linguistics, Peking University\\ 
  \textsuperscript{\rm 4}College of Science, Minzu University of China\\
  {\tt qingxiudong@icloud.com,\{wanxiaojun,yuecao\}@pku.edu.cn}}

\date{}

\begin{document}
\maketitle

\begin{abstract}
We propose ParaSCI, the first large-scale paraphrase dataset in the scientific field, including 33,981 paraphrase pairs from ACL (ParaSCI-ACL) and 316,063 pairs from arXiv (ParaSCI-arXiv). Digging into characteristics and common patterns of scientific papers, we construct this dataset though intra-paper and inter-paper methods, such as collecting citations to the same paper or aggregating definitions by scientific terms. To take advantage of sentences paraphrased partially, we put up PDBERT as a general paraphrase discovering method. The major advantages of paraphrases in ParaSCI lie in the prominent length and textual diversity, which is complementary to existing paraphrase datasets. ParaSCI obtains satisfactory results on human evaluation and downstream tasks, especially long paraphrase generation. 
\end{abstract}

\section{Introduction}

A paraphrase is a restatement of meaning with different expressions \cite*{bhagat2013paraphrase}. Being very common in our daily language expressions, it can also be applied to multiple downstream tasks of natural language processing (NLP), such as generating diverse text or adding richness to a chatbot. 

At present, paraphrase recognition or paraphrase generation are largely limited to the deficiency of paraphrase corpus. Especially, due to the permanent vacancy of paraphrase corpus in the scientific field, scientific paraphrase generation advances slowly. Scientific paraphrases can not only be helpful for data augmentation of challenging scientific machine translation, but is also effective for polishing scientific papers. However, existing paraphrase datasets are mainly from news, novels, or social media platforms. Most of them remain short sentences and interrogative or oral style. As a result, none of such training data can train out a scientific paraphrase generator. 
Taking the sentence \textit{“we used pos tags predicted by the stanford pos tagger”} as an example, the generated sentences from Transformer \cite{vaswani2017attention} models trained on existing paraphrase datasets\footnote{Here we use Quora Question Pairs and MSCOCO, they are introduced in Section \ref{s2}} are \textit{“level basic topics : what is the basic purpose of stanford traditional hmo”} and \textit{“a picture of a street sign with a sign on it”}, far from ground-truth paraphrases.
% \begin{table*}[htb]
% \centering
% \begin{tabular}{c|l}
% \hline
% \bf Training Data & \bf Generated Paraphrase\\
% \hline
% Quora & level basic topics : what is the basic purpose of stanford traditional hmo?\\
% MSCOCO & a picture of a street sign with a sign on it\\
% ParaSCI& for pos tagging , we used the stanford pos tagger.\\
% \hline
% \end{tabular}
% \caption{\label{font-table} Paraphrase Generated from Transformer models trained on different trainging data. The original sentence is “we used pos tags predicted by the stanford pos tagger.”. }
% \end{table*}

We have noticed that the structure of scientific papers is nearly fixed. Paraphrase sentence pairs appear not only within a paper (intra-paper) but also across different papers (inter-paper), which makes it possible to construct a paraphrase dataset in the scientific field. For example, repetitions of the same crucial contribution in a paper or explanations of the same term in different papers are potential paraphrases. Based on such characteristics, we design different methods to extract paraphrase pairs (shown in Section \ref{sec:method}).

In terms of the construction methods, existing methods merely focus on the paraphrase relationship between entire sentences, while hardly handle sentences with partial paraphrase parts, leaving much original corpus idle. We find that if part of a sentence paraphrases another short sentence, such sequences will be filtered out because the overall semantic similarity is not high enough. For paraphrase discovering in this case, we fine-tune BERT to extract semantically equivalent parts of two sentences and name it PDBERT. In order to train PDBERT, we construct pseudo training data by stitching existing paraphrase sentences, and train a paraphrase extraction model using the pseudo training data. In the end, this model performs well on real scientific texts.

After filtering, we obtain 350,044 paraphrase pairs and name this dataset ParaSCI. It consists of two parts: ParaSCI-ACL (33,981 pairs) and ParaSCI-arXiv (316,063 pairs). Compared with other paraphrase datasets, sentences in ParaSCI are longer and more sententially divergent. ParaSCI can be used for training paraphrase generation models. Furthermore, we hope that it can be applied to enlarge training data for other NLP tasks in the scientific domain.

% In a word, in view of the characteristics of scientific papers and the vacancy of paraphrase corpus in this field, we construct the first paraphrase corpus in the scientific field, which provides sufficient data for subsequent tasks related to scientific expression. 
Our main contributions include:
\begin{enumerate}
\item We propose the first large-scale paraphrase dataset in the scientific field (ParaSCI), including 33,981 pairs in ParaSCI-ACL and 316,063 pairs in ParaSCI-arXiv. Our dataset has been released to the public\footnote{https://github.com/dqxiu/ParaSCI}. 

\item We propose a general method for paraphrase discovering. By fine-tuning BERT innovatively, our PDBERT can extract paraphrase pairs from partially paraphrased sentences.
% It is also a general method for extracting paraphrase candidates extraction in other text.

\item The model trained on ParaSCI can generate longer paraphrases, and sentences are enriched with scientific knowledge, such as terms and abbreviations.
% provides paraphrase generation in scientific style and achieve good results (shown in Section \ref{sec:pg}).
% Implementing our ParaSCI dataset to Transformer baseline, we generate paraphrase sentences with scientific characteristics. It can be used as a baseline for subsequent paraphrase generation tasks in the scientific field.
\end{enumerate}
\section{Related Work}\label{s2}
Paraphrases capture the essence of language diversity \cite{pavlick2015ppdb} and play significant roles in many challenging NLP tasks, such as question answering \cite{dong2017learning}, semantic parsing \cite*{su2017cross} and machine translation \cite{cho2014learning}. Development in paraphrases relies heavily on the construction of paraphrase datasets. 
\paragraph{Paraphrase Identification Datasets}\citet*{dolan2005automatically} proposed MSR Paraphrase Corpus [MSRP], a paraphrase dataset of 5,801 sentence pairs, by clustering news articles with an SVM classifier and human annotations. 
% Besides, it has a known deficiency skewed toward over-identification \cite{socher2011dynamisocher2011dynamicc}, containing a large portion of sentence pairs with many ngrams shared in common and little variation.
As is discovered, platforms such as Twitter also contain many paraphrase pairs. Twitter Paraphrase Corpus [PIT-2015] \cite{xu2015semeval} contains 14,035 paraphrase pairs on more than 400 distinct topics. Two years later, Twitter Url Corpus [TUC] \cite{lan2017continuously} was proposed as a development of PIT-2015. TUC contains 51,524 sentence pairs, collected from Twitter by linking tweets through shared URLs and do not leverage any classifier or human intervention. 
Datasets such as MSRP or PIT-2015 encourage a series of work in paraphrase identification \cite{das2009paraphrase,mallinson2017paraphrasing} but the size limitation hinders complex generation models. 
\paragraph{Paraphrase Generation Datasets}
MSCOCO \cite{lin2014microsoft} was originally described as a large-scale object detection dataset.
% , now it is also used to evaluate paraphrase generation methods \cite{prakash2016neural}.
It contains human-annotated captions of over 120K images, and each image is associated with five captions from five different annotators. In most cases, annotators describe the most prominent object/action in an image, which makes this dataset suitable for paraphrase-related tasks. Consequently, MSCOCO makes great contribution to paraphrase generation. Quora released a new dataset\footnote{website:https://data.quora.com/First-Quora-Dataset-Release-Question-Pairs} in January 2017, which consists of over 400K lines of potential question duplicate pairs. 
% Each line contains IDS for each question in the pair, the full text for each question, and a binary value that indicates whether the questions in the pair are truly a duplicate of each other.
% It is also widely used in paraphrase generation, however, merely limited to short question generation.
\citet*{wieting-gimpel-2018-paranmt} constructed ParaNMT-50M, a dataset of more than 50 million paraphrase pairs. The pairs were generated automatically by translating the non-English side of a large parallel corpus. 
% ParaNMT-50M is a valuable resource for paraphrase generation. However, it is mainly from novels and contains a great portion of dialogues, resulting in a short average length as well.
Nowadays, MSCOCO and Quora are mainly used for paraphrase generation \cite{fu2019paraphrase,gupta2018deep}. Nevertheless, their sentence lengths or related domains are restricted. 
% In supplementary to existing paraphrase datasets, our ParaSCI focuses on the scientific field and possesses longer sentences. ParaSCI can provide more scientific knowledge and scientific information to paraphrase generation.
\section{Dataset}
\subsection{Source Materials}
Our ParaSCI dataset is constructed based on the following source materials:
\paragraph{ACL Anthology Sentence Corpus (AASC)} AASC \cite{aasc2018} is a corpus of natural language text extracted from scientific papers. It contains 2,339,195 sentences from 44,481 PDF-format papers from the ACL Anthology, a comprehensive scientific paper repository on computational linguistics and NLP.

\paragraph{ArXiv Bulk Data} ArXiv\footnote{website:https://arxiv.org/help/bulk\_data} is an open-access repository of electronic preprints. It consists of scientific papers in the fields of mathematics, physics, astronomy, etc.. As the complete set is too large to process, we randomly select 202,125 PDF files as our original data and convert them to TXT files, arranged by sentence.

\paragraph{Semantic Scholar Open Research Corpus (S2ORC)} S2ORC \cite{lo-wang-2020-s2orc} is a large contextual citation graph of scientific papers from multiple scientific domains, consisting of 81.1M papers, 380.5M citation edges. We select all the citation edges of ACL and arXiv from S2ORC for subsequent processing.

\subsection{Basic Information}
According to the source materials, ParaSCI includes two subsets, ParaSCI-ACL and ParaSCI-arXiv. Paraphrase pairs in ParaSCI-ACL focus on the NLP field, while paraphrase pairs in ParaSCI-arXiv are more general. Some cases are shown in Table \ref{tb1}. ParaSCI show three main highlights: 1) Sentences included are long, nearly 19 words per sentence; 2) Sentences are more sententially divergent; 3) It provides rich scientific knowledge.

\begin{table}[htb]
\centering
\small
\begin{tabular}{m{9mm}m{28mm}m{28mm}}
\toprule
\textbf{Name} & \textbf{Sentence A} & \textbf{Sentence B} \\ 
\midrule
ParaSCI-ACL& Word sense disambiguation (wsd) is the task of identifying the correct meaning of a word in context. & The process of identifying the correct meaning, or sense of a word in context, is known as word sense disambiguation (wsd).\\ 
\\
ParaSCI-ACL& In this paper, we study the use of standard continuous representations for words to generate translation rules for infrequent phrases. & In this work, we show how simple continuous representations of phrases can be successfully used to induce translation rules for infrequent phrases.\\
\\
ParaSCI-arXiv & Simon and Ronder propose a constellation model to localize parts of objects, which utilizes cnn to find the constellations of neural activation patterns. & Simon et al. propose a neural activation constellations part model to localize parts with constellation model. \\
\\
ParaSCI-arXiv & Here we will concentrate only on those aspects that are directly relevant to the odderon. & We will put some emphasis on those aspects that are immediately relevant to the odderon. \\
\bottomrule
\end{tabular}
\caption{\label{tb1} Example paraphrase pairs in ParaSCI. Sentence A and corresponding Sentence B are paraphrase pairs. }
\end{table}

\subsection{Statistic Characteristic}
To assess the characteristics of ParaSCI, we compare its statistic characteristics with five main sentential paraphrase datasets in Table \ref{tb2}. The source genre of ParaSCI is scientific papers. Therefore, sentences are more formal and scholastic, and they differ from oral TUC or newsy MSRP. The average sentence length of ParaSCI is almost twice as long as that of ParaNMT-50M, MSCOCO and Quora, and also much longer than that of TUC. Its average length is only a little shorter than that of MSRP. As MSRP only contains 3,900 gold-standard paraphrases, our ParaSCI is complementary to the vacancy of large-scale long paraphrase pairs.

The degree of alteration is another important aspect of paraphrases. To compare our ParaSCI with other existing paraphrase datasets in this aspect, we propose to calculate the BLEU4 score \cite{papineni2002BLEU} between the source and target sentences of each paraphrase pair and name it \textbf{Self-BLEU}. In Table \ref{tb2}, ParaSCI, especially ParaSCI-ACL, shows a relatively low Self-BLEU, which means sentences are significantly changed.

\begin{table*}[htb]
\centering
\setlength{\tabcolsep}{1.5mm}{
\begin{tabular}{llrrrrc}
\toprule
\textbf{Name} &\textbf{Genre}& \textbf{Size (pairs)} & \textbf{Gold Size$\tablefootnote{gold-standard paraphrase}$ (pairs)} & \textbf{Len} & \textbf{Char Len}&\textbf{Self-BLEU}\\ 
\midrule
MSRP  & news& 5,801& 3,900  & 22.48 & 119.62& 47.98 \\
TUC & Twitter& 56,787 & 21,287&15.55 & 85.10 & 12.53\\
ParaNMT-50M &Novels, laws&  51,409,585 & 51,409,585& 12.94 & 59.18& 28.60\\
MSCOCO   & Description& 493,186 & 493,186 & 10.48 & 51.56 & 31.97\\
Quora & Question&  404,289  & 149,263 &11.14  & 52.89 & 29.46 \\
ParaSCI-ACL  &Scientific Papers & 59,402 & 33,981 & 19.10 & 113.76& 26.52\\
ParaSCI-arXiv &Scientific Papers & 479,526 & 316,063 & 18.84 & 114.46& 29.90\\
\bottomrule
\end{tabular}}
\caption{\label{tb2} Statistic characteristics of main existing paraphrase datasets and our ParaSCI. As ParaNMT-50M is too large, we sample 500,000 pairs as representatives. Len means the average number of words per sentence and Char Len represents the average number of characters per sentence. We calculate Len, Char Len and Self-BLEU of the gold-standard paraphrases rather than the whole size of sentences. }
\end{table*}

\section{Method}\label{sec:method}
\subsection{Extracting Paraphrase Candidates}
Based on the unique characteristics of scientific papers, we extract the paraphrase sentences from a same paper (intra-paper) and across different papers (inter-paper). In most cases, we develop a simple but practical model to discover paraphrases in different sections effectively. For a more challenging case, when sentences are paraphrased partially rather than entirely, we propose PDBERT to collect more paraphrase candidates.

\subsubsection{Intra-paper Extraction of Paraphrase Candidates}
Authors usually write down the same information with transformed expressions repeatedly in different parts of the paper to emphasize critical information or echo back and forth. This kind of feature is the premise of our intra-paper extraction methods.

\paragraph{Sentence BERT for Paraphrase Extraction across Different Sections}\label{para:bert}

Noting that sentences with shared semantics appear in different parts of one paper.  For instance, the following sentences are from different parts of a same paper, and they are paraphrases:

$S_{1}$: \textit{we propose a simple yet robust stochastic answer network (SAN) that simulates multi-step reasoning in machine reading comprehension.} (\textit{\textbf{abstract}}, \citet{liu-etal-2018-stochastic})

$S_{2}$: \textit{we introduce Stochastic Answer Networks (SAN), a simple yet robust model for machine reading comprehension.} (\textit{\textbf{introduction}}, \citet{liu-etal-2018-stochastic})

However, sentences in \emph{Method}, \emph{Data} and \emph{Result} sections are semantically different even when they only have minor changes.  For example, the strings other than numbers may be very similar when presenting the two experimental results, but the semantics are completely different. Therefore, we mainly focus on six sections (\emph{Abstract}, \emph{Introduction}, \emph{Background}, \emph{Discussion}, \emph{Preamble} and \emph{Conclusion}). We directly obtain embeddings of sentences through BERT \cite{devlin2018bert}. Then, we calculate the cosine similarity pair by pair and retain sentence pairs with a similarity score higher than 0.931 as favorable paraphrase candidates. 16,563 paraphrase candidates from ACL and 158,227 paraphrase candidates from arXiv are obtained efficiently in this way.

It is noteworthy that as information is hardly repeated in the same section, we exclude the comparison between sentences in the same section.

\paragraph{PDBERT for Paraphrase Discovering}

In fact, two sentences, even if they are not semantically equivalent, may share some parts with common semantics. For example:

$S_{1}$: \textit{\textbf{rationales are never given during training}.}

$S_{2}$: \textit{in other words, target \textbf{rationales are never provided during training}; the intermediate step of rationale generation is guided only by the two desiderata discussed above.}

$S_{1}$ and the bold part of $S_{2}$ constitute a paraphrase. However, since the similarity between the entire $S_{1}$ and $S_{2}$ is only 0.88, this pair of sentences will be filtered out. This phenomenon frequently occurs when a sentence is much longer than the other, so that part of the long sentence paraphrases the short sentence. To resolve this problem, we propose a new paraphrase discovering model, PDBERT.

\begin{figure*}[!ht] 
\center{\includegraphics[width=13cm]  {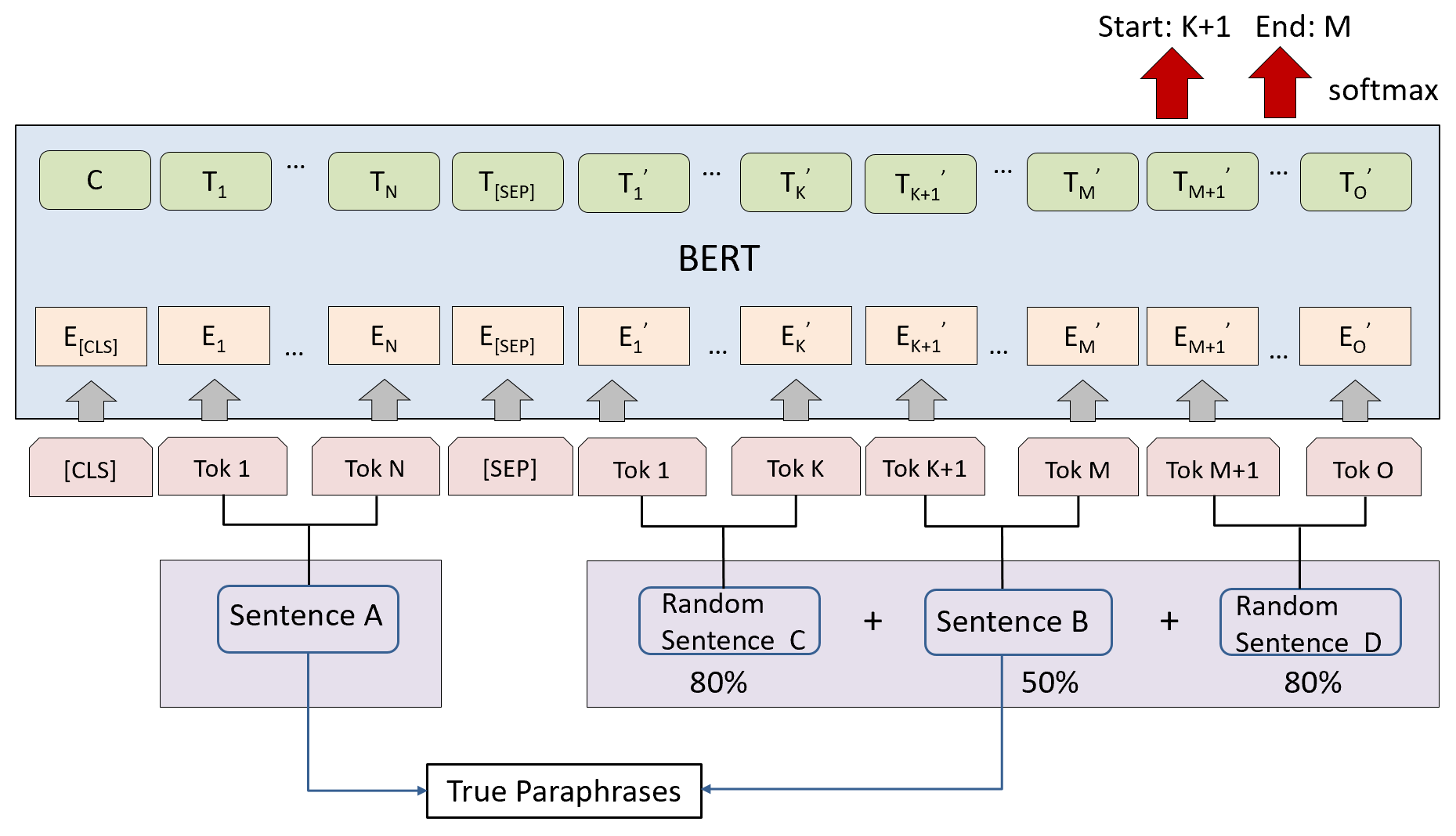}} 
\caption{A general paraphrase discovering model, PDBERT. We fine-tune a 12-layer BERT model for predicting the start and end positions of the paraphrase part. The bottom of the figure shows the construction of labeled data.}\label{f1} %图片名称和图片标号
\end{figure*} %结束
Our model architecture is shown in Figure \ref{f1}. We use the paraphrase sentence pairs recognized based on sentence similarity (our first method mentioned above) to construct pseudo-labeled data for model training. For each genuine paraphrase pairs, we take one as Sentence A and the other as Sentence B. Then we randomly pick out two sentences from the whole sentence set, as Sentence C and Sentence D. We use Sentence A as input 1, and concatenate Sentence C, Sentence B and Sentence D, with the 80\%, 50\%, and 80\% appearance probabilities respectively, as input 2. Input 1 and input 2 are further concatenated by adding a token [CLS] at the beginning of input 1, and inserting a [SEP] token between input 1 and input 2. The start and end positions of Sentence B in the concatenated string are recorded as the ground-truth. It is worth mentioning that while concatenating, the ending punctuations of Sentence C and Sentence B are removed. Besides, the random selection of sentences in input 2 guarantees our model to cover various situations, for example, the first part of input 2 is the ground-truth paraphrase of input 1, or input 2 is just Sentence B.

In this way, we generate a great number of training pairs. Input 1 represents a short sentence and input 2 represents a long sentence, which includes the paraphrase of the short one. We fine-tune the model to predict the start and end positions of Sentence B using a softmax function. 

Although our training data are pseudo, the fine-tuned model performs well on the real data. For a given real sentence pair, we take the shorter sentence as input 1 and the longer one as input 2, and extract from the long sentence according to the predicted positions. Actually, there is a tiny probability that the entire long sentence is the paraphrase of the short sentence. We exclude such a result because we have already obtained it via sentence BERT. Here we obtain 9,915 paraphrase candidates from ACL and 45,397 pairs from arXiv.

We compare our PDBERT model with a best-clause baseline. The latter is simply splitting the long sentence into several clauses by punctuation and selecting the combination of clauses that have a maximum similarity with the short sentence.
\begin{table}[ht]
% \small
\centering
\begin{tabular}{lrcc}
\toprule
\textbf{Method}& \textbf{Speed} & \textbf{Size} & \textbf{BERTScore} \\
\midrule
Best-clause&7.16& 341 &74.06\\
PDBERT&32.34& 387 &88.23\\
\bottomrule
\end{tabular}
\caption{\label{tb3}Comparison of paraphrase discovering methods. We divide the number of all the processed sentence pairs by time (720 minutes) as processing speed. Filtering out what can be obtained via sentence BERT, we take the number of remaining pairs as size of valuable extractions and use BERTScore (scibert-scivocab-uncased) to evaluate their quality.}
\end{table}
We implement PDBERT and the baseline method on the same corpus and compare the processing speed, size of valuable extractions and BERTScore\cite{zhang2019bertscore}. Table \ref{tb3} demonstrates the comparison. PDBERT has a higher extraction speed because, in the baseline method, we have to embed each possible combination of clauses. For valuable extractions, PDBERT extracts a larger size of candidate paraphrase pairs. Moreover, the BERTScore of PDBERT's results is higher, indicating its semantic advantage in paraphrase discovering.
\subsubsection{Inter-paper Extraction of Paraphrase Candidates}

Paraphrases also exist across different papers in the same field, including explanations of the same concept in different papers and citations to the same paper.

\paragraph{Explanations of the Same Concept}

In scientific papers, in order to introduce the definition of a task or a scientific terminology, the authors often explain it in one sentence. Therefore, various definitions of the same term in different papers become paraphrase candidates naturally, just as the following case:

$S_{1}$: \textit{\textbf{Sentence compression} is the task of producing a shorter form of a single given sentence, so that the new form is grammatical and retains the most important information of the original one.}

$S_{2}$: \textit{\textbf{Sentence compression} is a task of creating a short grammatical sentence by removing extraneous words or phrases from an original sentence while preserving its meaning.}

In order to extract the definition sentences from different papers, we design a series of possible patterns (regular expressions) of definition sentences and tag the terms in them. In the same subject or area (provided by meta-data of source materials), the extracted sentences are aggregated according to terms, so we obtain multiple explanations of the same concept. In order to ensure the same semantics, we combine them into pairs and adopt the method in Section \ref{para:bert} to filter out the sentence pairs that are semantically different. Here we get 5,912 paraphrase candidates from ACL and 63,258 paraphrase candidates from arXiv.

\paragraph{Citations to the Same Paper}

\begin{figure}[!ht] 
\flushleft 
\center{\includegraphics[width=75mm]  {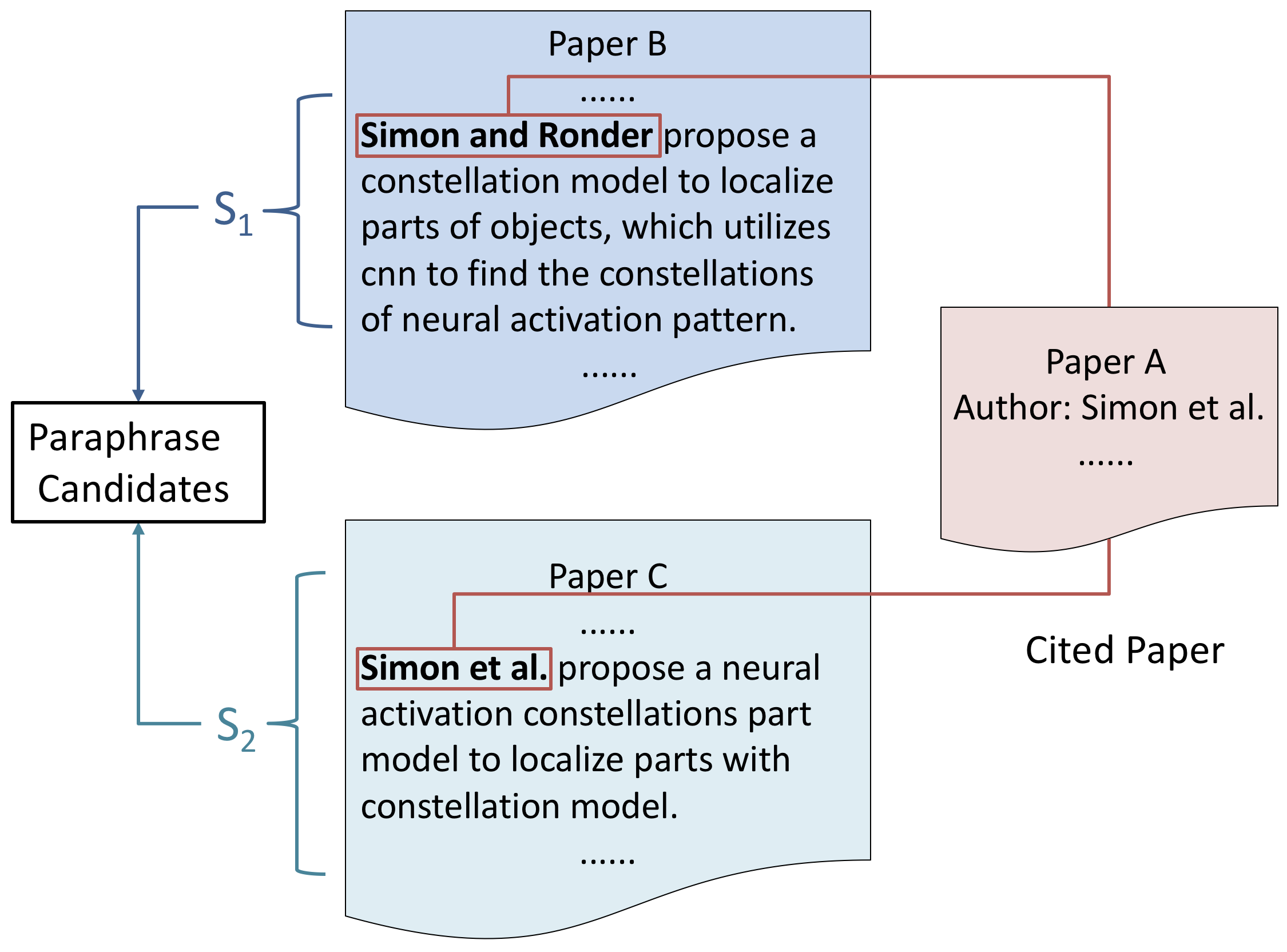}} 
\caption{Example of extracting paraphrase pairs from papers sharing a same cited paper.}\label{f2} %图片名称和图片标号
\end{figure}

Scientific papers often need to cite previous works. Besides, authors tend to give a brief introduction (i.e., citation text) to the cited paper. If different papers cite the same one, the introductory sentences to this cited paper in different papers also naturally constitute a sentence set with possible paraphrase relationships. Figure \ref{f2} provides an example.
 %结束
% \textit{$S_{1}$: \textbf{Collobert et al} used word embeddings as the input of various NLP tasks, including part-of-speech tagging, chunking, NER, and semantic role labeling.}

% \textit{$S_{2}$: In particular, \textbf{Collobert et al} and Turian et al learn word embeddings to improve the performance of in-domain POS tagging, named entity recognition, chunking and semantic role labelling.}

We locate the citations in the paper through S2ORC data and extract the citation sentence and the cited article. All the extracted results are aggregated according to the same cited paper. Then we match sentences in the same group. Similarly, in order to ensure the same semantics, we use the method in Section \ref{para:bert} to filter out semantically inconsistent sentence pairs. In this way, we obtain 27,016 pairs of candidates for ACL and 212,644 pairs for arXiv.

\subsection{Selecting High-quality Paraphrases}

Due to insufficient computing resources, we have used a rough but fast filtering method to construct paraphrase candidate set as mentioned above. It includes 59,406 pairs from ACL and 479,526 pairs from arXiv. To obtain the high-quality paraphrase corpus, we implement domain-related BERTScore and paraphrase length rates to determine if a candidate pair is really paraphrase. Our two-stage construction process works in a coarse-to-fine way. 

BERTScore leverages the pre-trained contextual embeddings from BERT and matches words in candidate and reference sentences by cosine similarity. It has been shown to correlate with human judgment on sentence-level and system-level evaluation. We calculate domain-related BERTScore for each paraphrase candidate, with the concrete setting of scibert-scivocab-uncased\_L8\_no-idf\_version=0.3.3.

We design paraphrase length rate (PLR) as another filter because the numbers of words in two paraphrase sentences usually do not vary too much. PLR is simply calculated as:

$$\frac{|L_A-L_B|}{\min (L_A,L_B)}$$\par
$L_A$ and $L_B$ stand for lengths of the corresponding sentences.

For paraphrase candidates extracted from explanations of the same concept, they consist of more abstract knowledge so we set a loose restriction. We select those with a BERTScore higher than 0.6 and PLR lower than 2. Therefore, we get 4,566 definition paraphrase pairs from ACL candidates and 49,052 pairs from arXiv candidates. For paraphrase candidates extracted from other methods, we change the threshold of BERTScore to be 0.7 and PLR to be 1.0. In this way, we get another 29,415 pairs from ACL candidates and 267,106 pairs from arXiv candidates.

\section{Manual Evaluation}
We conduct a manual analysis of our dataset in order to quantify its semantic consistency and literal variation lexically, phrasally and sententially. We employ 12 volunteers who are proficient in English to rate the instances. Three human judgements are obtained for every sample and the final scores are averaged across different judges.

\paragraph{Consistency Evaluation Criterion}For semantic consistency of paraphrase pairs, we design 5 degrees to distinguish. For a sentence pair to have a rating of 5, the sentences must have exactly the same meaning with all the same details. To have a rating of 4, the sentences are mostly equivalent, but some unimportant details can differ. To have a rating of 3, the sentences are roughly equivalent, with some important information missing or that differs slightly. For a rating of 2, the sentences are not equivalent, even if they share minor details. For a rating of 1, the sentences are totally different. (Examples are shown in the appendix)

\paragraph{Variation Evaluation Criterion}For literal variation of paraphrase pairs lexically, phrasally and sententially, we use the following criterion respectively: 5 means there are more than five variations of this level, 4 means four or five, 3 means two or three, 2 means it has only one change and 1 means no change.
% \begin{table}[]
% \centering
% \begin{tabular}{c|c}
% \hline
% & semantic consistency\\

% \hline
% ParaSCI-ACL & 4.17 \\
% ParaSCI-arXiv & 3.94 \\
% \hline
% \end{tabular}
% \caption{\label{font-table} Overall accuracy of ParaSCI from manual evaluation. }
% \end{table}
\begin{table}[]
\centering
\setlength{\tabcolsep}{1.0mm}{
\begin{tabular}{lccc}
\toprule
\textbf{Name}&\textbf{Lexical} & \textbf{Phrasal} & \textbf{Sentential}  \\
\midrule
ParaSCI-ACL & 3.82 & 2.73 & 2.01 \\
ParaSCI-arXiv & 3.67 & 2.68 & 1.48\\
\bottomrule
\end{tabular}}
\caption{\label{tb4} Overall variation of ParaSCI from manual evaluation. }
\end{table}
\paragraph{Quality Control} We evaluate the annotation quality of each worker using Cohen’s kappa agreement \cite*{artstein2008inter} against the majority vote of other workers. We asked the best worker to label more data by republishing the questions done by workers with low reliability (Cohen’s kappa \textless 0.4). Finally, the average Cohen’s kappa of semantic consistency evaluation is 0.71 and that of literal variation is 0.62 (0.66 lexically, 0.59 phrasally and 0.61 sententially).
\paragraph{Evaluation Results}The average semantic consistency of ParaSCI-ACL is 4.17 and that of ParaSCI-arXiv is 3.94, both around 4, which means most paraphrase pairs are nearly equivalent, only some unimportant details may differ. Besides, the average semantic consistency of ParaSCI-ACL is higher than that of ParaSCI-arXiv. In terms of literal variation, Table \ref{tb4} summarizes the annotations. ParaSCI-ACL and ParaSCI-arXiv show similar distributions. Paraphrase sentences usually change a lot lexically, because lexical variation is easier to realize. Although sentential variation scores are lower than lexical or phrasal scores, nearly one or two sentential variations for each pair are already rare and valuable for a paraphrase dataset. The long average length makes such sentential transformation possible, which is complementary to other datasets of short paraphrases. 

\section{Paraphrase Phenomenon Occurrence}

\begin{figure}[!ht] 
\flushleft 
\center{\includegraphics[width=75mm]  {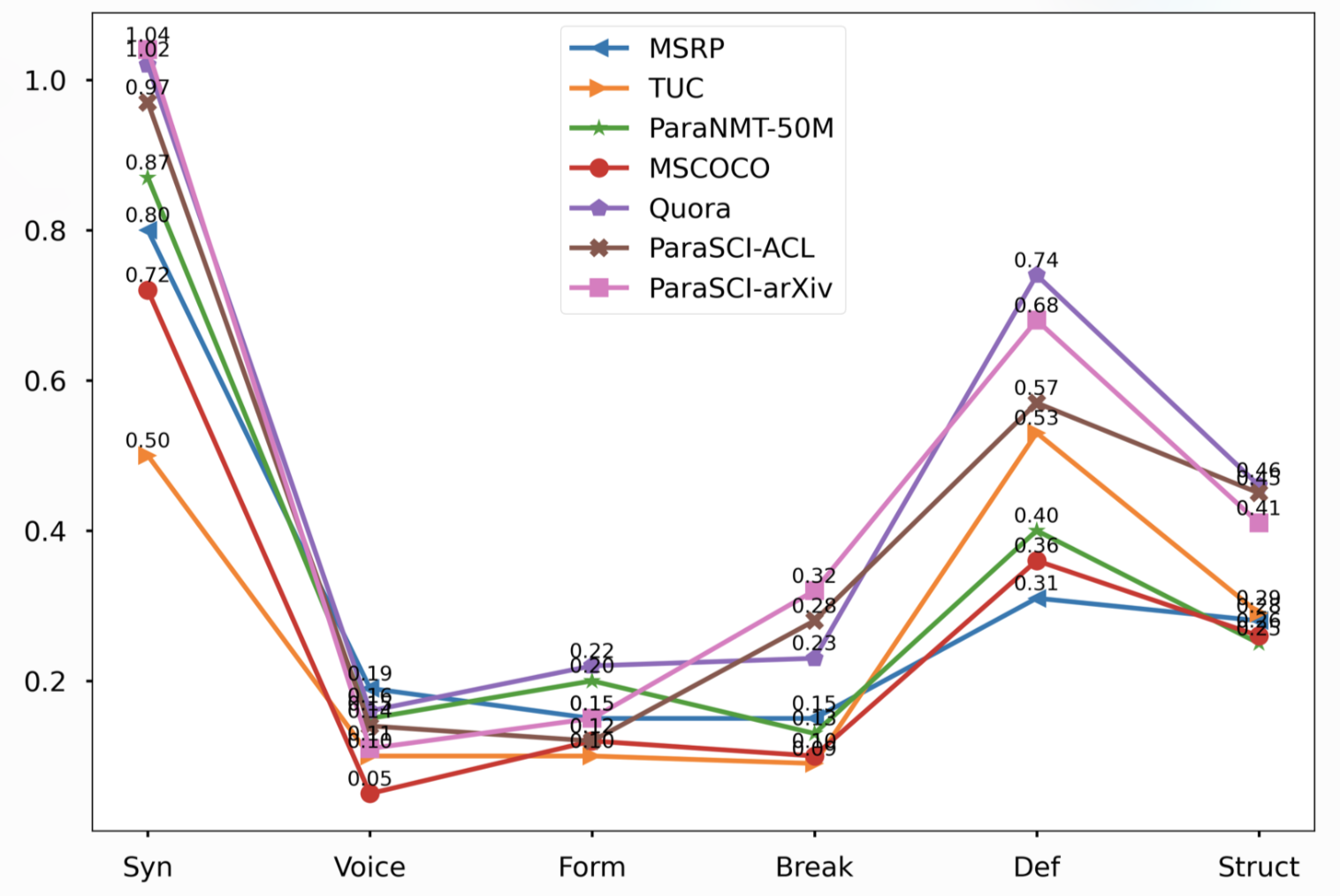}} 
\caption{Visualization of Table 5.}\label{f3} %图片名称和图片标号
\end{figure}

In order to show the differences across paraphrase datasets, we sample 100 sentential paraphrases from each dataset and count occurrences of each phenomenon. \citet*{boonthum-2004-istart} grouped common paraphrase phenomenon into 6 categories : \textbf{Synonym} (substitute a word with its synonym), \textbf{Voice} (change the voice of sentence from active to passive or vice versa), \textbf{Word-Form} (change a word into a different form), \textbf{Break} (break a long sentence down into small sentences), \textbf{Definition} (substitute a word with its definition or meaning), \textbf{Structure} (use different sentence structures to express the same thing).
We report the average number of occurrences of each paraphrase type per sentence pair for each corpus in Table \ref{tb5} and visualize that in Figure \ref{f3}.

\begin{table}[]
\centering
\small
\setlength{\tabcolsep}{0.9mm}{
\begin{tabular}{lcccccc}
\toprule
\textbf{Name}& \textbf{Syn} & \textbf{Voice} & \textbf{Form} & \textbf{Break}& \textbf{Def} & \textbf{Struct}\\
\midrule
MSRP & 0.80 & 0.19 & 0.15 & 0.15 & 0.31 & 0.28 \\
TUC & 0.50 & 0.10 & 0.10 & 0.09 & 0.53 & 0.29 \\
ParaNMT-50M & 0.87 & 0.15 & 0.20 & 0.13 & 0.40 & 0.25 \\
MSCOCO & 0.72 & 0.05 & 0.12 & 0.10 & 0.36 & 0.26 \\
Quora & 1.02 & 0.16 & 0.22 & 0.23 & 0.74 & 0.46 \\
ParaSCI-ACL & 0.97 & 0.14 & 0.12 & 0.28 & 0.57 & 0.45 \\
ParaSCI-arXiv & 1.04 & 0.11 & 0.15 & 0.32 & 0.68 & 0.41\\
\bottomrule
\end{tabular}}
\caption{\label{tb5} Example of extracting paraphrase pairs from papers sharing a same cited paper.}
\end{table}

 %结束

As shown in Table \ref{tb5} and Figure \ref{f3}, ParaSCI-ACL, ParaSCI-arXiv and Quora share a similar paraphrase phenomenon distribution. Besides, with respect to four categories of paraphrase phenomena (Synonym, Break, Definition, Structure), they all rank top 3. It reveals that our ParaSCI and Quora datasets contain more valuable paraphrase phenomena or paraphrase patterns. However, the Quora dataset is limited to questions pairs and our ParaSCI provides declarative sentence pairs. 

\section{Paraphrase Generation}\label{sec:pg}
To demonstrate further application of ParaSCI, we train paraphrase generation models on Quora, MSCOCO and ParaSCI. We then test their generation ability on the same scientific corpus, including sentences from ACL and arXiv respectively. For paraphrase generation from ParaSCI-ACL, we use 20,388 pairs for training, 6,796 pairs for validation and 6,797 pairs for test. For paraphrase generation from ParaSCI-arXiv, we use 189,639 pairs for training, 63,212 pairs for validation and 63,121 pairs for test. The BLEU scores and average lengths of generated sentences are shown in Table \ref{tb6}.

\begin{table}[t!]
\begin{center}
\setlength{\tabcolsep}{1.2mm}{
\begin{tabular}{llcc} 
\toprule
{\bf Training} & \bf Test & \bf BLEU & \bf Len \\
\midrule
Quora & ParaSCI-ACL & 6.31 & 14.23 \\
Quora & ParaSCI-arXiv & 8.06 & 13.92\\
ParaSCI-ACL & ParaSCI-ACL & 15.70 & 18.45\\
ParaSCI-arXiv & ParaSCI-arXiv & 27.18 & 18.82 \\
\bottomrule
\end{tabular}}
\end{center}
\caption{\label{tb6} BLEU scores and average lengths of scientific paraphrase generation by Transformer models trained on Quora and ParaSCI. Since the performance of the model trained on MSCOCO is rather poor (BLEU4 \textless 1.0), we omit the comparison with it. }
\end{table} 

Although there are many technology-related or scientific questions on Quora, the paraphrase generation model trained on Quora still fails to perform well in the scientific field, with low BLEU scores and short average length.
On the contrary, the paraphrase generation model trained on ParaSCI keeps generating longer sentences. The BLEU scores also demonstrate that the quality of the generated sentences is higher. This reflects the significant value of ParaSCI on paraphrase generation.

We show the generated paraphrases on different datasets in Table \ref{tb7}. To be fair, we still use the same Transformer architecture in the experiment. The generated paraphrases vary a lot. In most situations, the generated sentences from MSCOCO is incomplete, more like a phrase. The model trained on Quora only generates short questions. Whether trained on MSCOCO or Quora, the generated sentences usually share similar structures and have a large portion of entities in them. On the contrary, models trained on ParaSCI handle the paraphrase generation of longer sequence, including more modifiers and conjunctions.

Apart from that, generation models trained on ParaSCI manifest another characteristic. As ParaSCI consists of quantities of scientific terms and expressions, generation models trained on ParaSCI bring valuable scientific knowledge to the output sentence.

One thing to mention is that some abbreviations can be understood and utilized in the generation process. For instance, we generated \textit{“we ran \textbf{mt} experiments using the moses phrase-based translation system.”} for the sentence \textit{“we used moses as the phrase-based \textbf{machine translation} system.”}. As we use domain-related terms and corresponding abbreviations in scientific texts frequently, this advantage can add conciseness, technicality and naturality to the generated sentences.

Another thing to mention is that some common sense or scientific knowledge is also taken into paraphrase generation. For example, we input \textit{“the penn discourse treebank is the largest corpus richly annotated with explicit and implicit discourse relations and their senses.”} as the original sentence. It introduces the Penn Discourse Treebank \cite{miltsakaki2004penn} without any information related to its size and source, but the information is added to the generated sentences: \textit{“the penn discourse treebank is the largest available annotated corpora of discourse relations \textbf{over 2,312 wall street journal articles}.”}

These cases reveal different aspects of ParaSCI's advantages in paraphrase generation. In further work, we hope that ParaSCI will contribute more to scientific paraphrase generation and subsequently, be applied to more downstream tasks in the scientific field.

\begin{table}[ht]
\centering
\small
\begin{tabular}{m{10mm}m{28mm}m{28mm}}
\toprule 
\textbf{Name}& \textbf{Original} & \textbf{Paraphrase} \\ 
\midrule
MSCOCO & a group of people watch a dog ride a motorcycle.& an old photo of people riding on a motorcycle and waving.\\
\\
Quora & how can i get saved wifi password?&how can i see a saved wifi password?\\
\\

ParaSCI-ACL &relation extraction ( re ) is the task of determining semantic relations between entities mentioned in text.&relation extraction ( re ) is the task of recognizing the assertion of a particular relationship between two or more entities in text.\\
\\
ParaSCI-arXiv & cosmic strings are linear topological defects that can form in the early universe as a result of symmetry-breaking phase transitions.&cosmic strings are one-dimensional massive objects, which may appear as topological defects at the spontaneous symmetry breaking in the early universe.\\
\bottomrule
\end{tabular}
\caption{\label{tb7} Example paraphrase sentences generated by the same Transformer model trained on different datasets. Different from Table \ref{tb6}, models here are trained with in-domain data, so the training data and testing data come from the same field.}
\end{table}

\section{Conclusion and Future Work}
In this paper, we describe the characteristics and construction process of ParaSCI, a large-scale paraphrase dataset in the scientific field. It shows favorable results in the either automatic or manual evaluation.

For future work, although we filter out more than 200 thousand paraphrase candidates to promise the quality of ParaSCI, most candidates include valuable paraphrase patterns lexically or phrasally. Therefore, more paraphrase patterns are remaining to be discovered. Similarly, compared to the bulk data on arXiv or other scientific websites, we only use the tip of an iceberg to construct this dataset, and we are expecting to implement the methods in other scientific domains. For instance, we can obtain a biomedical paraphrase dataset from PubMed.

We hope that ParaSCI can be used to augment training data for various NLP tasks, such as machine translation in scientific field, and make more contributions to the development of NLP.

\section*{Acknowledgments}
This work was supported by National Natural Science Foundation of China (61772036), Beijing
Academy of Artificial Intelligence (BAAI) and Key Laboratory of Science, Technology and Standard in Press Industry (Key Laboratory of Intelligent Press Media Technology). We appreciate the anonymous reviewers for their helpful comments. Xiaojun Wan is the corresponding author. 

% \newpage
%\input{eacl2021.bbl}
\bibliography{eacl2021}
\bibliographystyle{acl_natbib}

\appendix
\clearpage
\section{Appendices}
\label{sec:appendix}
Examples of semantic consistency evaluation are shown in the following table.
\begin{table}[ht]
\centering
\small
\begin{tabular}{m{31mm}m{31mm}m{4mm}}
\toprule 
\textbf{Sentence A}& \textbf{Sentence B} & \textbf{Score} \\ 
\midrule
Task-oriented dialog systems help users to achieve specific goals with natural language.&We use a set of 318 English function words from the scikit-learn package.&1\\
\\
End-to-end task-oriented dialog systems usually suffer from the challenge of incorporating knowledge bases.&Task-oriented dialog systems help users to achieve specific goals with natural language.&2\\
\\
Opinion mining has recently received considerable attentions.&Analysis has received much attention in recent years.&3\\
\\
We evaluated all agents on 57 Atari 2600 games from the arcade learning environment.&We evaluated EMDQN on the benchmark suite of 57 Atari 2600 games from the arcade learning environment.&4\\
\\
Here we will concentrate only on those aspects that are directly relevant to the odderon. & We will put some emphasis on those aspects that are immediately relevant to the odderon.&5\\
\bottomrule
\end{tabular}
\caption{Examples of semantic consistency evaluation.}
\end{table}

\end{document}